\documentclass[10pt,twocolumn,letterpaper]{article}

\usepackage[pagenumbers]{cvpr} 

\usepackage{soul}
\usepackage[table]{xcolor}
\usepackage{multirow}
\usepackage{dblfloatfix}

\definecolor{cvprblue}{rgb}{0.21,0.49,0.74}
\definecolor{LightGray}{gray}{0.9}
\usepackage[pagebackref,breaklinks,colorlinks,allcolors=cvprblue]{hyperref}

\newcommand{\MYhref}[3][magenta]{\href{#2}{\color{#1}{#3}}}

\def\paperID{6254} 
\def\confName{CVPR}
\def\confYear{2026}

\title{DINO-Detect: A Simple yet Effective Framework for Blur-Robust AI-Generated Image Detection}

\author{Jialiang Shen\textsuperscript{1}\thanks{ Equal Contributions.} $\quad$ Jiyang Zheng\textsuperscript{1,2 $*$} $\quad$ Yunqi Xue\textsuperscript{3 $*$}$\quad$ Huajie Chen\textsuperscript{4} $\quad$ Yu Yao\textsuperscript{1}\\ 
Hui Kang\textsuperscript{1} $\quad$ Ruiqi Liu\textsuperscript{5} $\quad$ Helin Gong\textsuperscript{3} $\quad$ Yang Yang\textsuperscript{3} $\quad$ Dadong Wang\textsuperscript{2} $\quad$ Tongliang Liu\textsuperscript{1} \\
\small{
\textsuperscript{1}Sydney AI Center, The University of Sydney $\quad$ \textsuperscript{2}CSIRO, Data61} \\
\small{\textsuperscript{3}Shanghai Jiao Tong University $\quad$ \textsuperscript{4}City University of Macau $\quad$ \textsuperscript{5}CASIA}\\
{\tt\small shenjial12345@gmail.com, jzhe5740@uni.sydney.edu.au, 12223444lkkk.sjtu.edu.cn}
}

\begin{document}
\maketitle
\begin{abstract}

With growing concerns over image authenticity and digital safety, the field of AI-generated image (AIGI) detection has progressed rapidly. Yet, most AIGI detectors still struggle under real-world degradations, particularly motion blur, which frequently occurs in handheld photography, fast motion, and compressed video. Such blur distorts fine textures and suppresses high-frequency artifacts, causing severe performance drops in real-world settings.
We address this limitation with a blur-robust AIGI detection framework based on teacher-student knowledge distillation. A high-capacity teacher (DINOv3), trained on clean (i.e., sharp) images, provides stable and semantically rich representations that serve as a reference for learning. By freezing the teacher to maintain its generalization ability, we distill its feature and logit responses from sharp images to a student trained on blurred counterparts, enabling the student to produce consistent representations under motion degradation. 
Extensive experiments benchmarks show that our method achieves state-of-the-art performance under both motion-blurred and clean conditions, demonstrating improved generalization and real-world applicability. Source codes will be released at: \MYhref{https://github.com/JiaLiangShen/Dino-Detect-for-blur-robust-AIGC-Detection}{Project Page}.
\end{abstract}    
\section{Introduction}
\label{sec:intro}


\begin{figure}[t]
  \centering
  \includegraphics[width=0.8\linewidth, page=1]{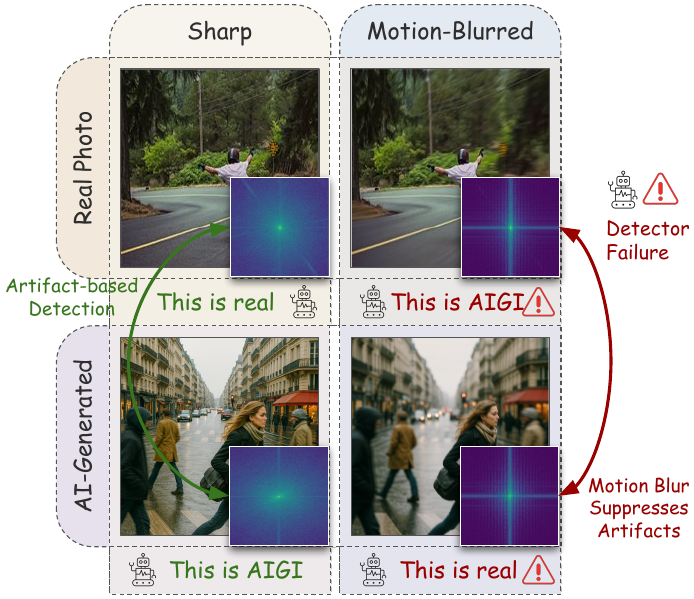}
  \caption{Motion blur suppresses discriminative high-frequency artifacts, leading to detector failure. In sharp images, AI-generated content contains excess mid–high-frequency energy from upsampling discontinuities, enabling artifact-based detectors to distinguish real from fake. When motion blur acts as a strong low-pass filter, these spectral cues vanish, collapsing the decision boundary and causing misclassification of both blurred reals and fakes.}
  \label{fig:att}
  \vskip -4mm
\end{figure}

With the rapid development of generative AI, particularly in image synthesis~\cite{kingma2013auto,goodfellow2014generative,ho2020denoising,lipman2022flow}, numerous powerful models~\cite{zhu2017unpaired,van2017neural,karras2019style,wang2019progan_training, zhang2023adding, peebles2023scalable,batifol2025flux, zheng2025aligning} have been introduced, capable of producing highly realistic and diverse images. While these advances have greatly expanded creative applications~\cite{rombach2022high, comanici2025gemini, labs2025flux1kontextflowmatching, zheng2024enhancing, zheng2025chainoffocus, tu2024ranked}, they have also intensified concerns regarding misinformation and content authenticity. Detecting AI-generated images (AIGIs) has therefore become an increasingly important task for ensuring media integrity and public trust~\cite{wu2023ai, cao2023comprehensive}.

Despite the remarkable progress in AIGI detection~\cite{lorenz2023detecting, yan2023ucf, zhong2023rich, wang2023dire, cozzolino2024raising, chen2024drct, yan2024sanity, liu2024forgery, lin2025seeing, zhou2025brought}, current detectors face a critical limitation when deployed in real-world environments: \emph{extreme vulnerability to motion blur}. In practical scenarios, images are frequently captured under dynamic conditions, such as handheld photography~\cite{liba2019handheld}, object movement~\cite{su2017deep}, or low-light scenes~\cite{hasinoff2016burst}, resulting in motion blur that degrades visual sharpness and suppresses high-frequency textures~\cite{sieberth2014motion, brooks2019learning, zhang2022deep}. Such degradations are unavoidable in real-world applications and are particularly common in online media, surveillance footage, and compressed videos, where motion blur naturally arises from acquisition or transmission processes. 

Figure~\ref{fig:att} compares the frequency energy distribution between real and fake images under sharp and blurred conditions. In the clear setting, fake images exhibit noticeably stronger responses in mid-to-high frequencies, a signature of generator-induced artifacts~\cite{tan2024frequency}. However, both natural and artificial blurs act as low-pass filters, suppressing these discriminative bands and flattening the radial spectra. The resulting spectral overlap between blurred real and fake images confirms that motion blur physically erases the high-frequency cues relied upon by artifact-based detectors~\cite{durall2020watch,wang2025spatial,liu2024forgery,shih2024does,liu2024npr, yan2024effort}, making genuine and synthetic images visually indistinguishable for existing detectors.

To address this critical issue, we propose \textit{DINO-Detect}, a \textbf{blur-robust general AIGI detector} designed to maintain reliable performance under real-world motion degradations. Our key idea is to leverage the semantic-level artifacts from the robust representations of a high-capacity vision transformer, DINOv3~\cite{simeoni2025dinov3}. Unlike conventional detectors that rely on high-frequency artifact patterns easily suppressed by motion blur, DINOv3 extracts global, semantically consistent representations invariant to low-level degradations. Its attention mechanism aggregates information across distant patches, enabling blur-invariant alignment between sharp and degraded views. Consequently, DINOv3 bridges the information gap induced by blur, preserving discriminative cues beyond pixel-level artifacts.

We further distill this generalization ability through a teacher-student knowledge-distillation framework~\cite{zhao2022decoupled}. The teacher, built upon a pretrained DINOv3 backbone, is further trained on sharp AIGI data to provide stable and well-generalized features as a reference for learning. Instead of directly fine-tuning the teacher on blurred data, which would distort DINOv3's pretrained representation space and weaken its generalization, we freeze its parameters and train a student detector to reproduce the teacher’s responses when given motion-blurred counterparts. This distillation process encourages the student to align its blurred-image embeddings with the teacher’s sharp-image features, effectively learning to generate consistent representations despite motion degradation.

To further strengthen blur invariance, we incorporate a contrastive loss~\cite{khosla2020supervised, zha2023rank, zheng2022towards} that explicitly aligns sharp-blurred pairs in the embedding space, reinforcing semantic consistency across degradation levels. This simple yet effective framework transfers the robustness of the teacher’s feature manifold to the student, enabling a detector to perform reliably under both clean and blurred conditions.

Extensive experiments demonstrate that \textit{DINO-Detect} achieves state-of-the-art performance under both motion-blurred and clean conditions. On our newly established \textit{AIGI-Blur Benchmark}, the first benchmark specifically designed to evaluate AIGI detectors under natural motion blur, our method improves accuracy by 10.27\% in realistic imaging scenarios. Moreover, on the existing \textit{general AIGCDetect Benchmarks}~\cite{zhong2023patchcraft, zhu2023genimage}, which contains clean AIGI data (i.e., sharp images) from a diverse set of generative models, our method further improves accuracy by an average of 1.59\% over the second-best performing detector. These results confirm that our blur-robust framework not only enhances robustness to real-world degradations, but also strengthens overall discriminative capability.

In summary, our main contributions are threefold:
(1) We identify and study the overlooked problem of motion blur in AIGI detection, revealing its severe impact on existing methods.
(2) We address the vulnerability of motion blur in the real-world setting by introducing \textit{DINO-Detect}, a teacher-student distillation framework that learns blur-invariant representations through clean-to-blur semantic alignment, enabling reliable detection across both degraded and pristine conditions.
(3) We establish the first motion-blur benchmark for AIGI detection and perform comprehensive evaluations demonstrating substantial improvements under both blurred and clean conditions.


\begin{figure*}[t]
  \centering
  \includegraphics[width=0.9\linewidth, page=1]{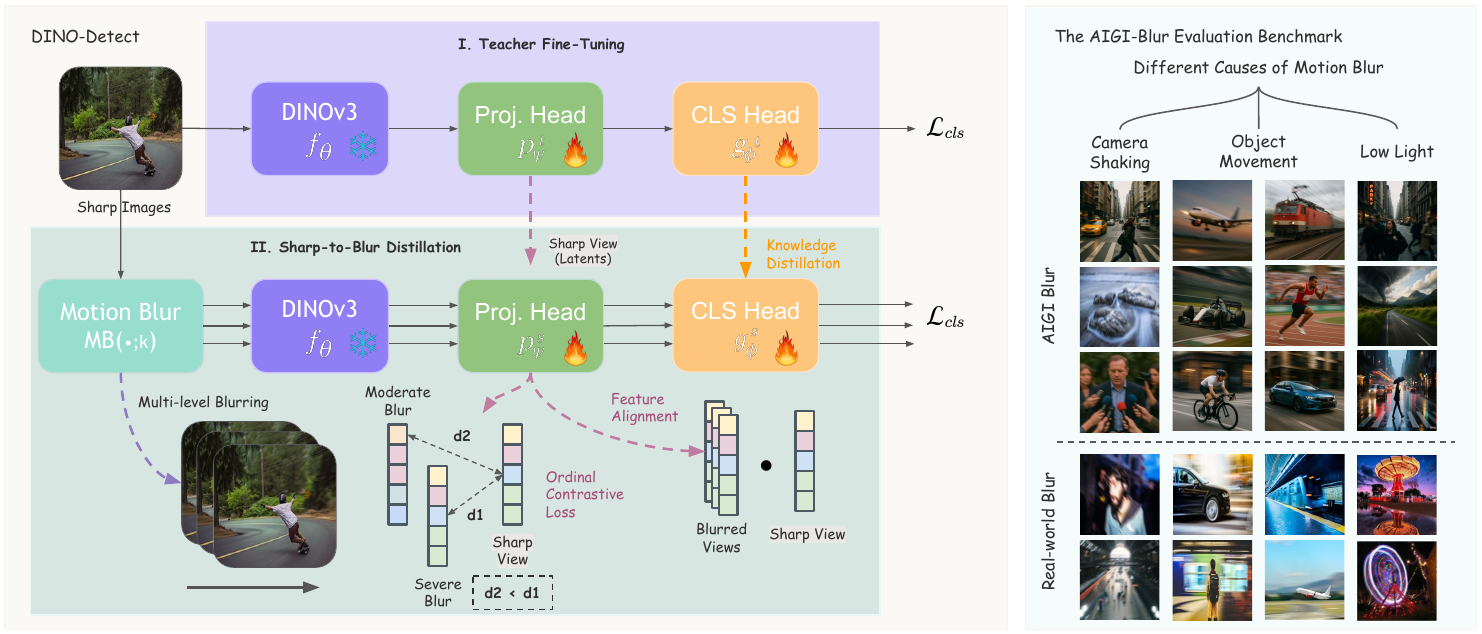}
  \caption{Illustration of our proposed DINO-Detect framework (Left) and the AIGI-blur evaluation benchmark (Right).}
  \label{fig:method}
  \vskip -3mm
\end{figure*}

\section{Related Works}
\label{sec:related}

Our work bridges three key areas: AI-generated content detection, adversarial robustness in computer vision, and motion blur analysis. AI-generated image detection has evolved from early frequency-domain and statistical analyses to more generalizable frameworks. Universal detection approaches such as \citet{ojha2023towards} leverage deep feature representations to capture artifacts common across generators, while \citet{park2025community} improve generalization through training on outputs from thousands of generators. Orthogonal subspace methods \citep{yan2024orthogonal} further disentangle generator-specific from universal features. Some other recent advances include attention-based detection mechanisms \citep{wang2023dire} that focus on discriminative regions and ensemble approaches \citep{li2022artifacts} that combine multiple detection strategies for enhanced robustness. More recently, \citet{zheng2024breaking} systematically evaluated AI-generated image detectors under real-world distortions, highlighting the vulnerability to degradations like blur. In the realm of physically grounded cues for synthetic content detection, \citet{jeon2025seeing} proposed a deepfake detection framework that leverages defocus blur an optically derived, depth-dependent phenomenon in real images to construct discriminative defocus maps, achieving strong performance on benchmarks like FaceForensics++~\cite{rossler2019faceforensics++} (average AUC 0.998), though their focus on defocus blur differs from the motion blur robustness gap addressed in our work. However, despite strong performance under standard conditions, the robustness of these detectors to real-world degradations like motion blur remains underexplored. 

The vulnerability of deep models to adversarial perturbations has been well studied \citep{goodfellow2014explaining,li2020face, chen2025stand, chen2025queen}, but most studies target imperceptible noise rather than natural degradations. Common corruptions such as blur significantly affect performance \citep{hendrycks2019benchmarking,zhang2024generative}, yet prior blur focusing works largely target face forgery detection and overlook underlying failure mechanisms. Motion blur, caused by relative motion between camera and scene, produces characteristic streaks \citep{chakrabarti2016neural} and alters attention patterns in vision transformers \citep{zamir2022restormer}. In this context, \citet{kim2024frequency} demonstrated that frequency-adaptive attention heads can recover fine-grained details under blur, but their effectiveness for synthetic content detection remains unproven. The impact of such blur-induced attention shifts on AI-generated detection has not been systematically studied, which directly motivates our work.

\section{Method}
\label{sec:method}
We address AI-generated image (AIGI) detection under real-world motion blur by coupling a strong DINOv3~\cite{simeoni2025dinov3} backbone encoder with a \emph{sharp-to-blur distillation} scheme. Given a labeled dataset
\(\mathcal{D}=\{(x_i,y_i)\}_{i=1}^N\), \(y_i\in\{0,1\}\), we synthesize paired views \(\big(x_i, x_i^{\mathrm{blur}}\big)\) via a physically grounded blur model and train light heads atop a frozen DINOv3 encoder. Training jointly enforces (i) supervised separability, (ii) sharp \(\leftrightarrow\) blur feature invariance, and (iii) guidance from sharp-view teacher logits. See Figure~\ref{fig:method}, Left.
\subsection{DINOv3 as the Feature Extractor}
\label{sec:dinov3}

Let \(f_\theta:\mathbb{R}^{H\times W\times 3}\!\to\!\mathbb{R}^{d}\) denote the pretrained DINOv3 ViT encoder. To preserve generalization, we freeze \(f_\theta\) by default. The encoded paired views are:
\begin{equation}
    h = f_\theta(x)\in\mathbb{R}^{d}, 
    \qquad
    h^{\mathrm{blur}} = f_\theta\!\big(x^{\mathrm{blur}}\big)\in\mathbb{R}^{d}.
\end{equation}
We add a two-layer projection head \(p_\psi:\mathbb{R}^{d}\!\to\!\mathbb{R}^{k}\) and a classifier \(g_\phi:\mathbb{R}^{k}\!\to\!\mathbb{R}^{2}\):
\begin{equation}
\begin{aligned}
    z = p_\psi^\mathrm{t}(h), \quad
    z^{\mathrm{blur}} = p_\psi^\mathrm{s}\!\big(h^{\mathrm{blur}}\big), \\
    u = g_\phi^\mathrm{t}(z), \quad
    u^{\mathrm{blur}} = g_\phi^\mathrm{s}\!\big(z^{\mathrm{blur}}\big).
\end{aligned}
\end{equation}
The head provides a space to impose invariance and contrastive structure while decoupling blur-robust shaping from the pretrained geometry. We normalize features where relevant, \(\hat{z}=z/\lVert z\rVert_2\). All \(p_\psi\) and \(g_\phi\) are trained from scratch.

\subsection{Sharp-to-Blur Distillation}
\label{sec:distill}
\paragraph{Blur model and paired sampling.}
\label{sec:blur_model}
For each sharp image \(x\), we synthesize a blurred counterpart \(x^{\mathrm{blur}}=\mathcal{B}(x;\kappa)\) by convolving with a motion point-spread function (PSF)~\cite{baxansky2014estimation} and composing with co-degradations that commonly co-occur with blur. We sample a hand-held camera-shake trajectory with length \(L\!\sim\!\mathrm{Uniform}(0,L_{\max})\) and global direction \(\alpha\!\sim\!\mathrm{Uniform}(0,\pi)\), with small angular jitter. The PSF is rasterized along the path and normalized to unit mass. We set \(L_{\max}\) to match deployment (e.g., \(L_{\max}\!\in\![7,21]\)). With probability \(p_d\), we apply defocus blur with kernel standard deviation \(\sigma\!\sim\!\mathrm{Uniform}(0,2.5)\). With small probabilities we apply JPEG compression (quality \(q\in[70,95]\)), mild sensor-like noise, and/or down–up sampling, preventing overfitting to a single kernel family.

\paragraph{Teacher fine-tuning on sharp views.}
We first construct a teacher model that learns reliable real-fake discriminative cues from sharp images. This design preserves DINOv3’s semantic priors while adapting them to the AIGI detection task, providing a stable reference for knowledge transfer. Specifically, the sharp-view teacher $(p_\psi^\mathrm{t}, g_\phi^\mathrm{t})$ is trained on clean images using a focal loss~\cite{lin2017focal} to handle class imbalance problem. Let $u=g_{\phi^\mathrm{t}}(p_{\psi^\mathrm{t}}(f_\theta(x)))$ and $p=\mathrm{softmax}(u)$. With class weights $\alpha$ and focusing parameter $\gamma$, the teacher objective is: 
\begin{equation}
\mathcal{L}_{cls}^{\mathrm{t}}
= - \sum_{c=1}^{C}\mathbf{1}[y{=}c]\;\alpha_c\;\big(1-p_c\big)^\gamma \,\log p_c.
\end{equation}
After convergence, the teacher's projection heads and classifier $(\psi^\mathrm{t},\phi^\mathrm{t})$ are then frozen for subsequent distillation.

\paragraph{Distillation objectives.}
\label{sec:student}
The student network is optimized to generalize across both sharp and blurred inputs by jointly enforcing accuracy and semantic alignment with the teacher. Specifically, it observes paired sharp-blur views of the same image and learns to (i) preserve feature similarity between the two views, and (ii) align the blurred prediction with the teacher’s sharp prediction. To realize these goals, four complementary objectives are combined: a focal loss for classification on blurred views $\mathcal{L}_{\mathrm{focal}}$:
\begin{equation}
    \mathcal{L}_{\mathrm{cls}} \;=\;
-\sum_{c=1}^{C}\mathbf{1}[y{=}c]\;\alpha_c\;\big(1-p^{\mathrm{blur}}_c\big)^\gamma \,\log p^{\mathrm{blur}}_c ,
        \label{eq:focal}
\end{equation}
where  $\alpha$ and $\gamma$ are the same class weights and focusing parameter as for teacher network. A feature-alignment loss $\mathcal{L}_{\mathrm{feat}}$ that encourages blur-invariant embeddings :
\begin{equation}
    \mathcal{L}_{\mathrm{feat}} \;=\;
         1 - \frac{1}{B}\sum_{i=1}^{B} \frac{\mathbf{f}_s^{(i)} \cdot \mathbf{f}_t^{(i)}}{\|\mathbf{f}_s^{(i)}\|_2 \|\mathbf{f}_t^{(i)}\|_2},
        \label{eq:feat}
\end{equation}
where $f_s$ and $f_t$ are the projected features of the student and teacher network. A knowledge-distillation loss~\cite{zhao2022decoupled} $\mathcal{L}_{\mathrm{KD}}$ is adapted to transfer semantic guidance from the teacher’s logits :
\begin{equation}
    \mathcal{L}_{\mathrm{KD}} \;=\;
        T^2\,\mathrm{KL}\!\left(
            \sigma\!\left(\tfrac{u}{T}\right)
            \,\middle\|\,
            \sigma\!\left(\tfrac{u^{\mathrm{blur}}}{T}\right)
        \right),
        \label{eq:kd}
\end{equation}
and lastly, an \emph{ordinal contrastive loss}~\cite{zha2023rank} $\mathcal{L}_{\mathrm{ordcon}}$ is employed to impose a monotonic ordering of sharp–blur distances in the representation space:
\begin{equation}
\begin{aligned}
    \mathcal{L}_{\mathrm{ordcon}}
    =
    &\frac{1}{N}
    \sum_{i=1}^{N}
    \frac{1}{N-1}
    \sum_{j\ne i}\\
    &-\log
    \frac{
        \exp\!\big(\mathrm{sim}(\hat{z}_i,\hat{z}_j)/\tau\big)
    }{
        \sum\limits_{\hat{z}_k:\,\Delta b_{i,k}\ge\Delta b_{i,j}}
        \exp\!\big(\mathrm{sim}(\hat{z}_i,\hat{z}_k)/\tau\big)
    },
\end{aligned}
    \label{eq:ordcon}
\end{equation}
where $\mathrm{sim}(\cdot,\cdot)$ denotes cosine similarity and $\Delta b_{i,j}$ represents the relative blur difference between samples. 
For a sharp anchor $\hat{z}_i$, views with milder blur (smaller $\Delta b_{i,j}$) are enforced to have higher similarity than those with stronger blur, thereby encoding blur severity as a continuous order in the embedding space. 
Together with the preceding terms, this encourages discriminative yet blur-consistent representations, yielding a detector that remains reliable under real-world degradations. The total loss is
\begin{equation}
\label{eq:total}
    \mathcal{L} \;=\;
\lambda_1\underbrace{\mathcal{L}_{\mathrm{cls}}}_{\text{task}} \hspace{0.1em}
+ \hspace{0.1em} \lambda_2 \hspace{0.1em} \smash{\underbrace{\mathcal{L}_{\mathrm{feat}}}_{\text{blur inv.}}}
+ \lambda_3\!\!\smash{\underbrace{\mathcal{L}_{\mathrm{KD}}}_{\text{sharp guide}}}
\!\!\!+ \lambda_4\hspace{0.1em}\smash{\underbrace{\mathcal{L}_{\mathrm{ordcon}}}_{\text{blur struct.}}}.
\end{equation}

\section{The AIGI-Blur Benchmark}
To rigorously evaluate model robustness under realistic motion degradations, we construct the AIGI-Blur Benchmark, a curated dataset comprising both AI-generated and real-world motion-blurred images across diverse scenarios. The benchmark is designed to capture the heterogeneity of motion blur encountered in everyday photography and video content, offering a challenging testbed for assessing the generalization of AIGI detectors beyond clean conditions.

\noindent
\textbf{AIGI motion-blur synthesis.} We first synthesize 3.5K AI-generated motion-blurred images using multiple state-of-the-art diffusion models, including Lumina 2.0~\cite{lumina2}, Stable Diffusion 1.5~\cite{Rombach_2022_CVPR}, Stable Diffusion 3~\cite{esser2024scaling}, FLUX~\cite{batifol2025flux}, and SDXL~\cite{podell2024sdxl}. Each model generates visually diverse content spanning natural scenes, objects, and human portraits to reflect the variety of generative outputs in real applications (See Figure~\ref{fig:method}, Right). Motion blur is simulated under three representative real-world conditions: 1) Camera shaking, which produces global streaks due to unsteady hand-held capture, 2) Object movement, which introduces spatially varying blur from fast-moving subjects, and 3) Low-light capture, where prolonged exposure exacerbates motion trails. The prompts (including blur descriptions) are created stochastically within physically plausible ranges to ensure perceptual realism while maintaining diversity across scenarios.

\noindent
\textbf{Real-world motion-blur collection.}
To complement the synthetic AIGI data, we gather an equal number of real-world motion-blurred images from open-source image repositories and publicly available video datasets. In addition to manually curating blurred photographs from professional and amateur photography collections, we extract frames from videos exhibiting dynamic object motion or rapid camera panning, where naturally occurring motion blur is evident. This hybrid strategy ensures that the real subset spans a broad spectrum of blur patterns and lighting conditions, aligning with the characteristics of real handheld and mobile-capture scenarios.

\noindent
\textbf{Benchmark characteristics.}
Each image in AIGI-Blur is annotated with its source type (AI-generated or real), blur scenario, and blur severity, enabling both binary detection and fine-grained robustness analysis. By incorporating both synthetically controlled and naturally observed motion blur, the benchmark facilitates systematic evaluation of detectors under progressively challenging degradation conditions. Together, these properties make AIGI-Blur a comprehensive and physically grounded benchmark for assessing the reliability of AIGI detectors in the wild.









\begin{table*}[t]
\centering
\caption{Performance comparison on GenImage benchmark
~\cite{zhu2023genimage}. Models are trained on SDv1.4 training set and evaluated on seven unseen generation methods. The best results are highlighted in \textbf{bold} and the second best are \underline{underlined}.}
\label{tab:genimage_results}
\resizebox{0.8\textwidth}{!}{
\begin{tabular}{l|cccccccc>{\columncolor{LightGray}}c}
	\toprule
	\textbf{Method} & 	\textbf{ADM} & 	\textbf{VQDM} & 	\textbf{Midjourney} & 	\textbf{Wukong} & 	\textbf{GLIDE} & 	\textbf{BigGAN} & \textbf{SDv1.4}&	\textbf{SDv1.5} & 	\textbf{Avg.} \\
\midrule
ResNet-50~\cite{he2016deep} & 53.50 & 56.60 & 54.90 & 98.20 & 61.90 & 52.00 & 99.90&99.70 & 72.10 \\
DeiT-S~\cite{touvron2021training} & 49.80 & 56.90 & 55.60 & 98.90 & 58.10 & 53.50 &99.90 &\underline{99.80}&  71.60 \\
Swin-T~\cite{liu2021swin} & 49.80 & 62.30 & 62.10 & 99.10 & 67.60 & 57.60 & 99.90&\underline{99.80}& 74.80 \\
CNNSpot~\cite{wang2020cnn}  & 50.10 & 53.40 & 52.80 & 78.60 & 39.80 & 46.80 & 96.30&95.90&  64.20 \\
Spec~\cite{zhang2019detecting} & 49.70 & 55.60 & 52.00 & 94.80 & 49.80 & 49.80 &99.40 &99.20& 68.80 \\
F3Net~\cite{qian2020thinking} & 49.90 & 49.90 & 50.10 & \textbf{99.90} & 50.00 & 49.90 &99.90 &\textbf{99.90} & 68.70 \\
GramNet~\cite{liu2020global} & 50.30 & 50.80 & 54.20 & 98.90 & 54.60 & 51.70 &99.20 &99.10 & 69.90 \\
UnivFD~\cite{ojha2023towards}  & 58.10 & 67.80 & 91.50 & 94.50 & 73.40 & 57.70 & 96.40&96.10 & 79.50 \\
NPR~\cite{tan2024rethinking} & 77.45 & 84.94 & 81.62 & 97.67 & 90.50 & 84.20 & 98.20&98.68 & 89.33 \\
FreqNet~\cite{tan2024frequency} & 67.27 & 76.39 & 90.30 & 98.07 & 87.20 & 81.40 & 98.80&99.38 & 87.57 \\
FatFormer~\cite{liu2024forgery} & 76.46 & 97.92 & 93.43 & 99.00 & 88.71 & 55.80 &\textbf{100.00} &99.00 & 88.93 \\
DRCT ~\cite{chen2024drct}& 79.40 & 90.00 & 91.50 & 94.70 & 89.20 & 81.70 & 95.00&94.40 & 89.50 \\
Effort~\cite{yan2024effort} & 78.70 & 91.70 & 82.40 & 97.40 & \underline{93.30}& 77.60 &99.80 &\underline{99.80} & 91.10 \\
AIDE~\cite{yan2024sanity} & 78.54 & 80.26 & 79.38 & 98.65 & 91.82 & 66.89 & 99.74 & 99.76& 86.88 \\
Dual-Data~\cite{chen2025dual} & \textbf{94.30}& \underline{94.60}&\textbf{94.40} &96.30 & 93.10 & \underline{95.80}& 97.30& 97.20& \underline{95.50}\\
\midrule
Ours & 	\underline{87.82} & 	\textbf{99.32} & 	\underline{93.85} &  \underline{99.60} & 	\textbf{97.39} & 	\textbf{98.94} &99.79 & 99.64 & \textbf{96.77} \\
\bottomrule
\end{tabular}
}
\end{table*}

\section{Experiments}
\label{sec:exp}

\subsection{Experimental Setup}
\subsubsection{Datasets}
Experiments are conducted on widely-adopted benchmarks for both in-domain evaluation and cross-domain evaluation on AIGI detection task. As well as the AIGI-Blur dataset.\\
\textbf{GenImage~\cite{zhu2023genimage}.} This dataset primarily employs diffusion-based models for generating images, including Midjourney, Stable Diffusion v1.4 (SDv1.4), SDv1.5, ADM, GLIDE, Wukong, VQDM, and BigGAN. Following the same experiment settings~\cite{zhu2023genimage,liu2024npr}, SDv1.4 is treated as the training set, and evaluating on all the generation methods. Given the diverse image sizes within GenImage, images smaller than 224 pixels are cropped to 224 pixels. \\
\textbf{AIGCDetectBenchmark~\cite{zhong2023aigcdetect}.} It includes a multi-category dataset that adopts a hierarchical structure where each semantic category (\eg, 	\textit{airplane}, 	\textit{bird}, 	\textit{car}) contains two subfolders: 	\texttt{0\_real} (human-captured images) and 	\texttt{1\_fake} (synthetically generated images). The ProGAN training split is utilized as the primary training corpus\cite{wang2019progan_training}. The test set encompasses diverse generative methods to evaluate cross-method generalization capability. The images in this dataset are cropped to 448 pixels during the training and testing procedure.

\subsubsection{Implementations}
\textbf{Backbone Architecture.} The pre-trained DinoV3-ViT-7B with patch size 16 $\times$ 16~\cite{simeoni2025dinov3} is adopted as the encoder for both the student and teacher in our framework. 
\textbf{Network Design.} The T-S network architecture consists of:
\begin{itemize}[leftmargin=*,noitemsep,topsep=2pt]
    \item 	\textit{Teacher model}: A multi-layer projection head (3 layers with hidden dimension 2048$\rightarrow$1024$\rightarrow$512) followed by a two-layer GELU-activated classifier. The projection dimension is set to 512.
    \item 	\textit{Student model}: The frozen DinoV3 backbone is shared with the teacher, but independent projection (2 layers, 4096$\rightarrow$1024$\rightarrow$512) and classification heads are maintained with a higher dropout rate (0.2 vs 0.1).
\end{itemize}
\textbf{Data Pre-processing.} As previously introduced, we employed different data processing methods for different training sets. Taking the ProGAN training set~\cite{simeoni2025dinov3} as an example, all input images are resized to $512\times512$ pixels and randomly cropped to $448	\times448$ during training. Test images are center-cropped to $512	\times512$. Pixel intensities are normalized using ImageNet statistics: $\mu=(0.485,0.456,0.406)$ and $\sigma=(0.229,0.224,0.225)$.\\
\textbf{Data Augmentation.} Different augmentation strategies are employed for teacher and student training phases:
\begin{itemize}[leftmargin=*,noitemsep,topsep=2pt]
    \item \textit{Teacher phase}: Strong augmentation is applied, including balanced ColorJitter (brightness/contrast/saturation$\pm$0.1, hue$\pm$0.05), random rotation ($\pm5^\circ$), and probabilistic JPEG compression (quality 85-95, $p=0.3$).
    \item 	\textit{Student phase}: In addition to teacher augmentations, probabilistic motion blur augmentation is applied using the blur model (see Sec.~\ref{sec:blur_model}). Three modes are supported, including: 1) Global blurring, 2) Category-conditional blurring (CCMBA)~\cite{rajagopalan2023improving}, and 3) Mixed blurring.
\end{itemize}
\begin{table*}[t]
\centering
\caption{Performance comparison on AIGCDetectBenchmark~\cite{zhong2023aigcdetect} (Part 1, GAN-based generators). All models are trained on ProGAN~\cite{wang2019progan_training} training set. The best results are in \textbf{bold}, and the second best are \underline{underlined}.}
\label{tab:aigc_results_gan}
\resizebox{0.85\textwidth}{!}{
\begin{tabular}{l|cccccccc>{\columncolor{LightGray}}c}
	\toprule
	\textbf{Method} & \textbf{ProGAN} & \textbf{StyleGAN} & \textbf{BigGAN} & \textbf{CycleGAN} & \textbf{StarGAN} & \textbf{GauGAN} & \textbf{StyleGAN2}  && \textbf{Avg.} \\
	\midrule
	CNNSpot~\cite{wang2020cnn} & \textbf{100.00} & 90.17 & 71.17 & 87.62 & 94.60 & 81.42 & 86.91 && 87.41\\
	FreDect~\cite{frank2020leveraging} & 99.36 & 78.02 & 81.97 & 78.77 & 94.62 & 80.57 & 66.19 && 82.79\\
	Fusing~\cite{ju2022fusing} & \textbf{100.00} & 85.20 & 77.40 & 87.00 & 97.00 & 77.00 & 83.30 && 86.70\\
	LNP~\cite{liu2022detecting} & 99.67 & 91.75 & 77.75 & 84.10 & \textbf{99.92} & 75.39 & 94.64 && 89.03\\
	LGrad~\cite{tan2023learning} & 99.83 & 91.08 & 85.62 & 86.94 & 99.27 & 78.46 & 85.32 && 89.50\\
	UnivFD~\cite{ojha2023towards} & 99.81 & 84.93 & 95.08 & \textbf{98.33} & 95.75 & \underline{99.47} & 74.96 && 92.62\\
	DIRE-G~\cite{wang2023dire} & 95.19 & 83.03 & 70.12 & 74.19 & 95.47 & 67.79 & 75.31 && 80.16\\
	DIRE-D~\cite{wang2023dire} & 52.75 & 51.31 & 49.70 & 49.58 & 46.72 & 51.23 & 51.72 && 50.43\\
	PatchCraft~\cite{zhong2023patchcraft} & \textbf{100.00} & 92.77 & \underline{95.80} & 70.17 & 99.97 & 71.58 & 89.55 && 88.55\\
	NPR~\cite{tan2024rethinking} & 99.79 & \underline{97.70} & 84.35 & 96.10 & 99.35 & 82.50 & \textbf{98.38} && \underline{94.02}\\
	AIDE~\cite{yan2024sanity} & \underline{99.99} & \textbf{99.64} & 83.95 & 96.20 & \underline{99.91} & 73.25 & \underline{98.00} && 92.99\\
	Dual-Data~\cite{chen2025dual} & 92.80 & 87.80 & 91.00 & 72.50 & 72.70 & 92.70 & 90.20 && 85.67\\
    \midrule
	Ours & \underline{99.99} & 93.41 & \textbf{99.08} & \underline{96.40} & 88.14 & \textbf{99.83} & 93.55 && \textbf{95.77}\\
	\bottomrule
\end{tabular}
}
\end{table*}

\begin{table*}[t]
\centering
\caption{Performance comparison on AIGCDetectBenchmark~\cite{zhong2023aigcdetect} (Part 2, Diffusion and other modern generators).The settings are identical as in Table~\ref{tab:aigc_results_gan}. The best results are in \textbf{bold}, and the second best are \underline{underlined}.}
\label{tab:aigc_results_diff}
\resizebox{0.8\textwidth}{!}{
\begin{tabular}{l|ccccccccc>{\columncolor{LightGray}}c}
	\toprule
	\textbf{Method}& \textbf{DALLE-2}&\textbf{WFIR} & \textbf{ADM} & \textbf{Glide} & \textbf{Midjourney} & \textbf{SDv1.4} & \textbf{SDv1.5} & \textbf{VQDM} & \textbf{Wukong} &   \textbf{Avg.} \\
	\midrule
	CNNSpot~\cite{wang2020cnn}&50.45 & \textbf{91.65} & 60.39 & 58.07 & 51.39 & 50.57 & 50.53 & 56.46 & 50.45 &  58.69\\
	FreDect~\cite{frank2020leveraging}&34.70& 50.75  & 63.42 & 54.13 & 45.87 & 38.79 & 39.21 & 77.80 & 34.70 &  50.58\\
	Fusing~\cite{ju2022fusing}& 52.80& 66.80 & 49.00 & 57.20 & 52.20 & 51.00 & 51.40 & 55.10 & 52.80 &  54.44\\
	LNP~\cite{liu2022detecting}& 88.75& 70.85  & 84.73 & 80.52 & 65.55 & 85.55 & 85.67 & 74.46 & 88.75 &  79.51\\
	LGrad~\cite{tan2023learning} & 65.45 & 55.70 & 67.15 & 66.11 & 65.35 & 63.02 & 63.67 & 72.99 & 65.45 &  64.93\\
	UnivFD~\cite{ojha2023towards}&50.00& 86.90  & 66.87 & 62.46 & 56.13 & 63.66 & 63.49 & 85.31 & 50.75 &  66.95\\
	DIRE-G~\cite{wang2023dire}& 66.48 & 58.05 & 75.78 & 71.75 & 58.01 & 49.74 & 49.83 & 53.68 & 66.48  & 60.41\\
	DIRE-D~\cite{wang2023dire}& 92.45 &53.30  & \textbf{98.25} & 92.42 & 89.45 & 91.24 & 91.63 & 91.90 & 92.45 & 87.58\\
	PatchCraft~\cite{zhong2023patchcraft}& \underline{96.60}&85.80  & 82.17 & 83.79 & \underline{90.12} & 95.38 & 95.30 & 88.91 & 96.60 &  89.76\\
	NPR~\cite{tan2024rethinking}& 20.00& 65.80  & 69.69 & 78.36 & 77.85 & 78.63 & 78.89 & 78.13 & 64.90 & 74.03\\
	AIDE~\cite{yan2024sanity}& \underline{96.60}& 84.20  & 93.43 & \underline{94.09} & 77.20 & 93.00 & 92.85 & \underline{95.16} & 96.60  & \underline{90.82}\\
	Dual-Data~\cite{chen2025dual}& \textbf{98.80} & 52.10  & 89.50 & 89.60 & \textbf{95.60} & \underline{98.70} & \textbf{98.60} & 76.60 & \textbf{98.80}  & 87.44\\
    \midrule
	Ours & 92.00& \underline{89.60} & \underline{96.79} & \textbf{94.19} & 67.50 & \textbf{98.79} & \underline{96.62} & \textbf{98.21} & \underline{97.87}  & \textbf{92.41}\\
	\bottomrule
\end{tabular}
}
\end{table*}
\noindent
\textbf{Training.}
Teacher training is conducted for 4 epochs using AdamW optimizer~\cite{loshchilov2017decoupled} with learning rate $\eta=1	\times10^{-4}$ and weight decay $\lambda=1	\times10^{-4}$. The learning rate follows a cosine annealing schedule. The hyperparameters for Focal Loss~\cite{lin2017focal} are $\alpha=1.0$ and $\gamma=2.0$. Batch size is set to 128 per GPU.
Student heads are optimized for 15 epochs. Loss coefficients are set to $(\lambda_{\text{cls}}, \lambda_{	\text{distill}}, \lambda_{\text{feature}}) = (1.0, 1.0, 0.5)$. Student optimization uses AdamW with learning rate $\eta=5	\times10^{-5}$, weight decay $\lambda=1	\times10^{-4}$, and cosine annealing schedule. Batch size is 128 per GPU.
\subsubsection{Evaluation Protocol}
\textbf{Metrics.} Classification accuracy are reported for both the student model on held-out test sets. Following standard practice~\cite{ojha2023fakedetect,liu2024npr}, the following evaluation protocols are conducted:
\begin{itemize}[leftmargin=*,noitemsep,topsep=2pt]
    \item 	\textit{Cross-dataset generalization}: Models are trained on one dataset (SDv1.4 or ProGAN) and tested on all categories of both benchmarks.
    \item 	\textit{Cross-method generalization}: Evaluation is performed on unseen generation methods within each benchmark.
\end{itemize}

\subsection{Main Results}
\paragraph{General AIGI Benchmarks.} As shown in~\cref{tab:genimage_results,tab:aigc_results_gan,tab:aigc_results_diff}, our method achieves consistent state-of-the-art performance across all major benchmarks and generation families. On the GenImage benchmark~\cite{zhu2023genimage}, which evaluates unseen diffusion-based generators, our model attains an average accuracy of 96.77\%, surpassing all previous methods, and outperform the second best state of the art detector by a margin of 1.27\%, and demonstrating strong generalization to diverse diffusion architectures (e.g., GLIDE~\cite{nichol2022glide}, Midjourney, and SD-v1~\cite{rombach2022high}). When tested on the AIGCDetectBenchmark (\cref{tab:aigc_results_gan}) containing GAN-based generators, our approach reaches 95.77\% average accuracy, outperforming previous leading detectors such as NPR - 94.02\% and AIDE - 92.99\%. This indicates that our framework preserves discriminative capability even when trained on a single source but tested across unseen generative styles. In the Diffusion and modern-generator setting (\cref{tab:aigc_results_diff}), our model again establishes a new best average of 92.45\%, substantially higher than the second-best: Dual-Data~\cite{chen2025dual} of 87.44\%), verifying strong resilience to large distribution shifts introduced by diffusion models and modern AIGIs. Notably, our detector consistently ranks top-1 on nearly every individual generator, particularly achieving 99.6-99.8\% accuracy on SD-v1.x and Wukong, confirming its robustness to both architectural diversity and visual degradations. Overall, these results validate the effectiveness of our proposed framework in achieving generalization and cross-model transferability.

\paragraph{Blur Robustness and Out-of-Domain Generalization.}
\cref{tab:wildrf_generalization,tab:wildrf_generalization_1,tab:synthbuster_results} evaluate the robustness of our detector under real-world motion blur and domain shifts. On the WildRF dataset~\cite{cavia2024real}, our method attains an average accuracy of 95.80\% on clean social media images and 86.67\% under blurred conditions, maintaining a remarkably small degradation of only 9.1 pp, while competing methods such as Effort~\cite{yan2024effort} and NPR~\cite{tan2024rethinking} experience drops exceeding 20 pp. This clearly demonstrates that our model generalizes effectively from synthetic training to real-world degradations. On the Chameleon benchmark~\cite{zeng2025chameleon}, which stresses cross-domain and blur robustness jointly, our student model achieves 93.68\% on clean images and 86.28\% under blur, surpassing all prior detectors. These consistent gains confirm that our blur-aware framework not only enhances motion-blur robustness but also improves real-world generalization to unseen degradations and domains.

\begin{table}[t]
\centering
\caption{Out-of-domain generalization on WildRF dataset. Models trained on GenImage (SDv1.4) are evaluated on real-world social media images from Facebook, Twitter, and Reddit. Results are reported on both clean and blurred versions.}
\label{tab:wildrf_generalization}
\resizebox{\columnwidth}{!}{
\begin{tabular}{l|ccc|ccc}
	\toprule
\multirow{2}{*}{\textbf{Method}} & \multicolumn{3}{c|}{	\textbf{Clean Images}} & \multicolumn{3}{c}{	\textbf{Blurred Images}} \\
\cmidrule(lr){2-4} \cmidrule(lr){5-7}
 & 	\textbf{Facebook} & 	\textbf{Twitter} & 	\textbf{Reddit} & 	\textbf{Facebook} & 	\textbf{Twitter} & 	\textbf{Reddit} \\
 CNNDet~\cite{wang2020cnn} & 70.60 & 71.40& 75.40 &60.00&64.20& 52.73\\
 UnivFD~\cite{ojha2023fakedetect} &71.56 & 69.72 & 79.60 &50.62 &52.73  &55.21 \\
 CLIP~\cite{radford2021learning} &  78.40 & 78.10 & 80.80 &55.31& 49.32& 45.33\\
AIDE-Progan~\cite{yan2024sanity}      &70.20  &69.16 &68.83 &49.69 &64.85 &52.73   \\
AIDE-SDv1.4~\cite{yan2024sanity}      &74.69  &72.27 &75.52 &49.06 &50.62 &56.60    \\
Effort-SDv1.4~\cite{yan2024effort}    &85.51 &82.19 & 77.93  & 67.50 & 74.35   &68.13 \\
Effort-AllGen~\cite{yan2024effort} &81.04 &82.70 &90.60 & 64.06& 58.33 & 63.00    \\
NPR~\cite{tan2024rethinking}& 76.60 & 79.50 & 89.80 & 59.69&56.03 &52.00 \\
 
\midrule

Ours & 	\textbf{93.81} & 	\textbf{97.15} & 	\textbf{95.80} & 	\textbf{89.38} & 	\textbf{89.14} & 	\textbf{86.67} \\
\bottomrule
\end{tabular}
}
\end{table}

\begin{table}[t]
\centering
\caption{Performance on the Chameleon benchmark. Results are reported on both clean and blurred versions to evaluate robustness. Models are trained on GenImage (SDv1.4) }
\label{tab:synthbuster_results}
\resizebox{\columnwidth}{!}{
\begin{tabular}{l|cc}
    \toprule
    \textbf{Method} & \textbf{Chameleon (Clean)} & \textbf{Chameleon (Blur)} \\
    \midrule
     CNNDet~\cite{wang2020cnn} & 63.14 & 55.75\\
      UnivFD~\cite{ojha2023fakedetect} &59.67 &51.69  \\
      CLIP~\cite{radford2021learning} & 77.98 & 63.69\\
      
    AIDE-Progan~\cite{yan2024sanity}         & 58.77 & 56.60    \\
    AIDE-SDv1.4~\cite{yan2024sanity}         & 62.60 & 51.81    \\
    
    Effort-SDv1.4~\cite{yan2024effort}       & 66.63 & 60.33    \\
    Effort-ALLGenimages~\cite{yan2024effort} & 76.07 & 67.51    \\
     NPR~\cite{tan2024rethinking}& 66.87 & 60.03 \\
    \midrule
    
    Ours                             & \textbf{93.68} & \textbf{86.28} \\
    \bottomrule
\end{tabular}
}
\end{table}

\begin{table}[t]
  \centering
  \caption{Comparison of detection accuracy on our newly constructed benchmark containing AI-generated and real images degraded by realistic motion blur. The benchmark evaluates model robustness under blur-induced artifact suppression. Our method achieves the highest performance, outperforming all existing detectors by a large margin and demonstrating superior generalization to blur-corrupted generative content.}
  \label{tab:baseline_performance_blur_dataset}
  \renewcommand{\arraystretch}{1.2}
  \setlength{\tabcolsep}{6pt}
  \resizebox{\columnwidth}{!}{
  \begin{tabular}{ccccccc}
    \toprule
     \textbf{CLIP}~\cite{radford2021learning} & \textbf{DRCT}~\cite{chen2024drct} & \textbf{AIDE}~\cite{yan2024sanity} & \textbf{UnivFD}~\cite{ojha2023towards} & \textbf{NPR}~\cite{tan2024rethinking} & \textbf{Effort}~\cite{yan2024effort} & \textbf{Ours} \\
    \midrule
    \rowcolor[HTML]{F8F8F8}
    61.16 & 54.72 & 68.27 & 42.72 & 44.33 & 70.76 & \textbf{81.03} \\
    \bottomrule
  \end{tabular}
  }
\end{table}
\paragraph{Motion-Blurred AIGI Benchmarks.}
Tab.~\ref{tab:baseline_performance_blur_dataset} reports the detection accuracy on our Motion-Blurred AIGI Benchmark, designed to assess model robustness under severe motion blur. Our method achieves the highest accuracy of 81.03\%, substantially surpassing the strongest existing baseline, Effort by over 10 percentage points. Classical artifact-based detectors such as NPR~\cite{liu2024npr} of 44.33\% and DRCT~\cite{chen2024drct} of 54.72\% perform poorly, confirming that blur effectively suppresses the high-frequency traces on which they rely. Representation-based models such as CLIP~\cite{radford2021learning} and UnivFD~\cite{ojha2023fakedetect} also struggle to maintain discriminative power when blur disrupts texture-level cues. In contrast, our Sharp-to-Blur Distillation framework preserves semantic and structural features crucial for robust detection, enabling stable generalization to blurred AI-generated content. This result validates the effectiveness of our approach in mitigating motion-blur degradation and highlights its strong resilience against real-world image distortions.

\paragraph{Attention Distortion under Motion Blur.}
\begin{figure}[t]
  \centering
  \caption{Impact of motion blur on model attention patterns. The plots show the average similarity between attention maps of clean and motion-blurred images across varying blur kernel sizes.}
  \vskip -2mm
  \includegraphics[width=0.98\linewidth, page=1]{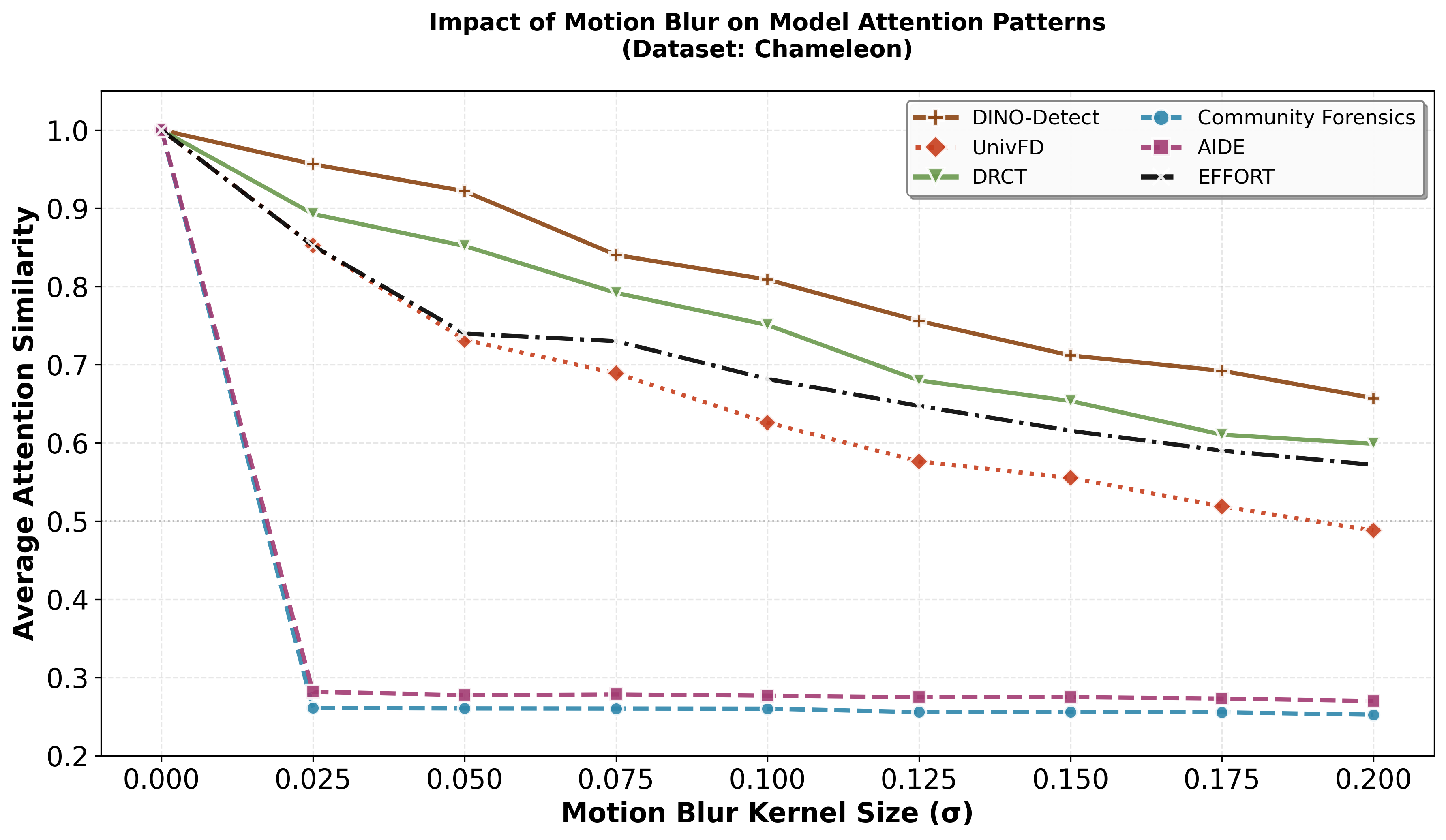}
  \label{fig:impact_wild}
  \vskip -4mm
\end{figure}
\begin{figure}[t]
  \centering
  \includegraphics[width=0.98\linewidth, page=1]{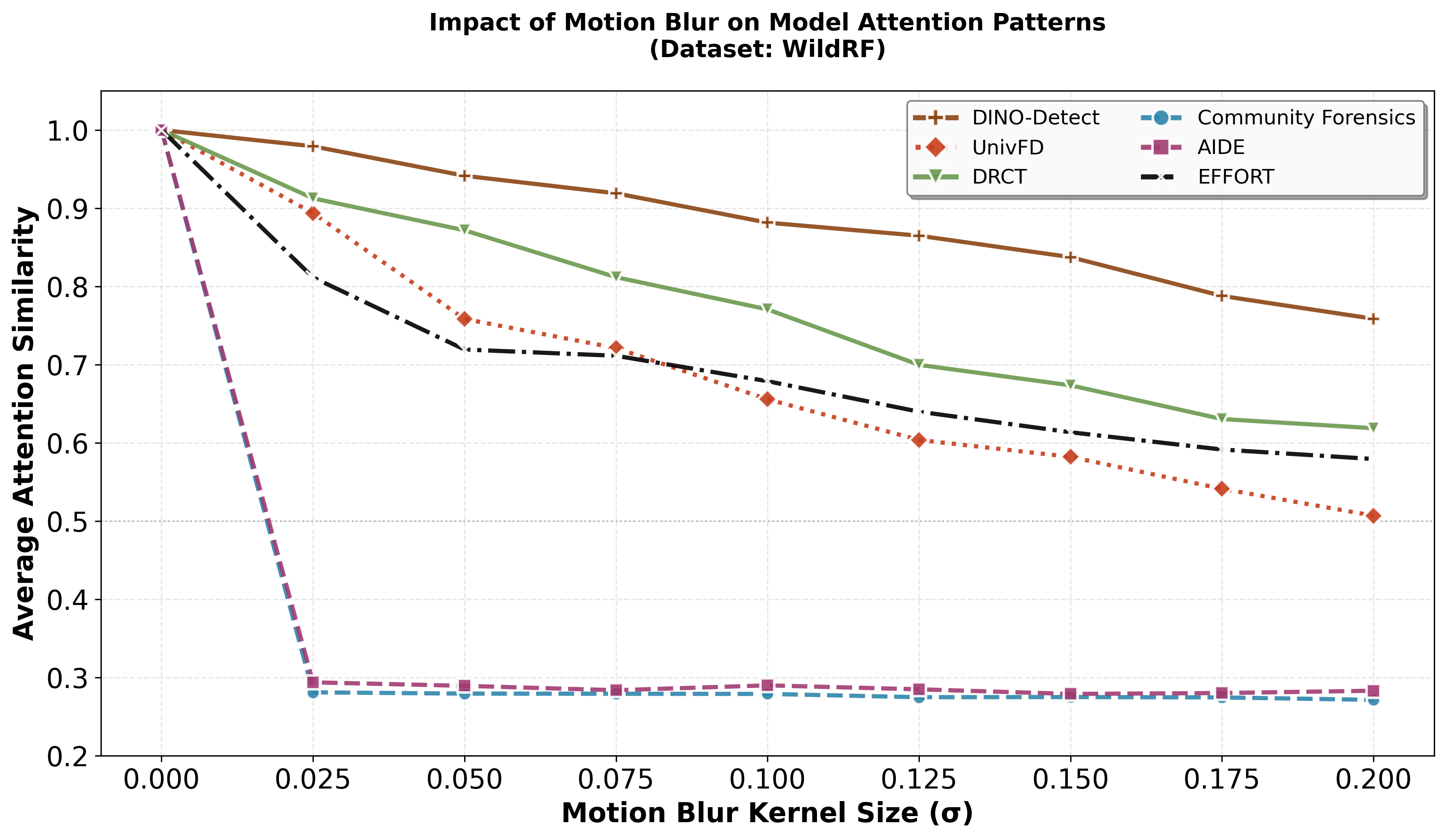}
  \label{fig:impact_chame}
  \vskip -6mm
\end{figure}
\begin{figure*}[t]
  \centering
  \caption{Patch-Level Structural Consistency under Motion Blur. We visualize the patch-wise similarity matrices of blurred fake (\textcolor{pink}{top row}) and real (\textcolor{LimeGreen}{bottom row}) images extracted by three models: CLIP-ViT~\cite{radford2021learning}, UnivFD~\cite{ojha2023towards}, and our DINO-Detect.}
  \vskip -2mm
  \includegraphics[width=\linewidth, page=1]{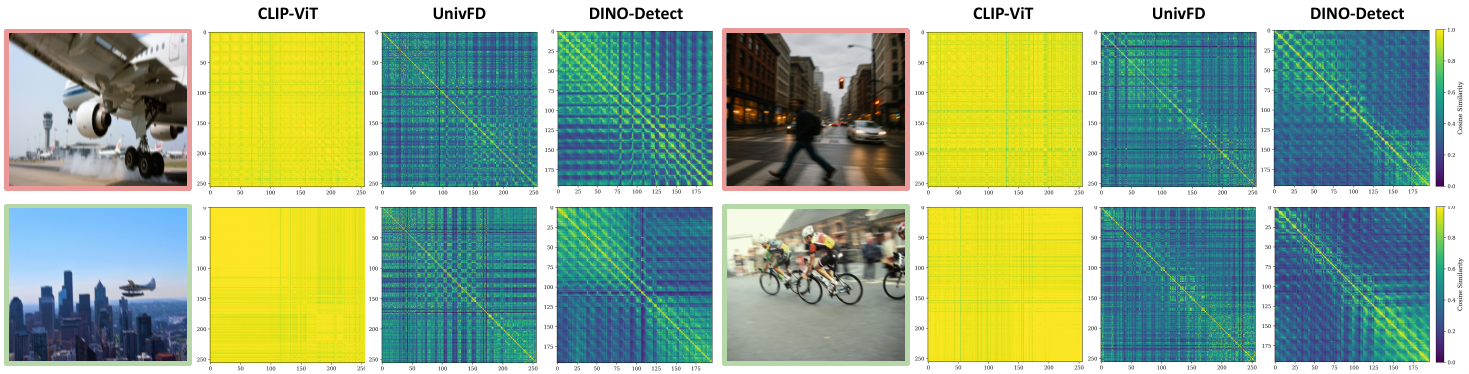}
  \label{fig:case}
  \vskip -6mm
\end{figure*}

Figure~\ref{fig:impact_wild} shows how motion blur affects the attention consistency of different detectors on the WildRF and Chameleon datasets. As the blur kernel size increases, all methods experience reduced attention similarity, yet the degree of degradation differs notably. Traditional forensic models (AIDE~\cite{yan2024sanity}, EFFORT~\cite{yan2024effort}, Community Forensics~\cite{park2025community}) collapse quickly, with similarity dropping below 0.3 even under mild blur, revealing heavy dependence on high-frequency cues. In contrast, representation-based models (DINO-Detect, UnivFD~\cite{ojha2023fakedetect}, DRCT~\cite{chen2024drct}) exhibit stronger stability, with our proposed DINO-Detect maintaining above 0.7 similarity under strong blur. These results highlight the attention distortion effect, blur shifts focus from semantic to background regions, and demonstrate the effectiveness of learned feature invariance in preserving structured attention under motion degradation.

\paragraph{Case Study on Structural Information Preservation.}
Figure~\ref{fig:case} visualizes the patch-wise similarity matrices of blurred fake and real images. The similarity matrices reveal each model’s ability to preserve structural information under motion blur. CLIP-ViT~\cite{radford2021learning} shows highly saturated and uniform activations, indicating poor spatial discrimination and loss of structural awareness once blur is introduced. UnivFD~\cite{ojha2023fakedetect} partially retains block-wise correlations but exhibits noisy and inconsistent patterns, suggesting that its representations are sensitive to blur-induced texture distortion. In contrast, DINO-Detect preserves clear diagonal and localized similarity structures in both fake and real cases, indicating a more coherent and geometry-aware representation. Notably, DINO-Detect maintains distinct internal correlations even for blurred fakes, reflecting its robustness to motion-induced degradation and its enhanced capacity to capture underlying object structure rather than superficial texture cues.

\paragraph{Blur-Type Robustness.}
To further investigate the effect of our blur-robust distillation, we evaluate the detector on four distinct blur types: Gaussian, Box, Radial, and Bokeh, using the WildRF dataset~\cite{cavia2024real} (As shown in~\cref{tab:wildrf_generalization_1}). Each blur simulates a different real-world degradation pattern, from isotropic smoothing to directional and defocus distortions. Our method achieves the highest accuracy across all categories, with 76.29\% on Gaussian blur, 79.10\% on Box blur, 83.38\% on Radial blur, and 77.83\% on Bokeh blur. In contrast, the best competing approach, Effort~\cite{yan2024effort} lags by more than 10 pp on average, confirming that existing detectors struggle under strong blur corruptions. The superior performance of our model across both spatially uniform and anisotropic blurs validates the effectiveness of our blur-aware distillation and feature-alignment design, which enable stable representation learning even when high-frequency cues are suppressed. This analysis highlights our framework not only generalizes well to unseen blur types but also learns to exploit structure and semantic-based cues that remain invariant under degradation.

\begin{table}[t]
\centering
\caption{Out-of-domain generalization with different blur model on WildRF dataset. Models trained on GenImage are evaluated on real-world social media images. Results are reported the average accuracy from different blurred types.}
\label{tab:wildrf_generalization_1}
\resizebox{\columnwidth}{!}{
\begin{tabular}{l|c|c|c|c}
\toprule
BlurType & \textbf{Gaussian blur} & \textbf{Box blur} & \textbf{Radial blur} & \textbf{Bokeh blur} \\
    \midrule
     CNNDet~\cite{wang2020cnn} & 62.25 &62.31  & 62.04 &61.88\\
     UnivFD~\cite{ojha2023fakedetect} &48.92 &52.18 &54.32 & 54.66\\
     CLIP~\cite{radford2021learning} & 50.88&61.18 &53.46 &50.98\\
    AIDE-Progan~\cite{yan2024sanity} &57.68 &57.84 &57.26 &57.36  \\
    AIDE-SDv1.4~\cite{yan2024sanity} & 55.76&58.87 &56.76 &57.96 \\
    Effort-SDv1.4~\cite{yan2024effort} &67.96 &64.56 &64.89 &67.15 \\
    Effort-AllGen~\cite{yan2024effort} &57.94 &57.96 &60.18 &61.36 \\
    NPR~\cite{tan2024rethinking}&59.96 &58.70 &57.51 &55.43 \\
    \midrule
    Ours & \textbf{76.29} & \textbf{79.10} & \textbf{83.38} & \textbf{77.83} \\
    \bottomrule
\end{tabular}
}
\end{table}

\section{Conclusions}
\label{sec:conclusions}
This work introduces a simple yet effective sharp-to-blur distillation framework that substantially enhances the robustness of AI-generated image detectors against motion blur. By leveraging the semantic stability of a frozen DINOv3 teacher and enforcing sharp–blur consistency through feature and logit distillation, our approach preserves discriminative representations even under severe degradations. The proposed design generalizes well across diverse datasets, setting new benchmarks for blur-robust AIGI detection and offering a practical path toward trustworthy media forensics in real-world environments.

{
    \small
    \bibliographystyle{ieeenat_fullname}
    \bibliography{main}
}


\end{document}